\PassOptionsToPackage{table}{xcolor}
\documentclass{bmvc2k}

\title{Pose-Robust Calibration Strategy for\\Point-of-Gaze Estimation on Mobile Phones}

\addauthor{Yujie Zhao}{zhaoyujie24s@ict.ac.cn}{1,2}
\addauthor{Jiabei Zeng}{jiabei.zeng@ict.ac.cn}{1,2}
\addauthor{Shiguang Shan}{sgshan@ict.ac.cn}{1,2}

\addinstitution{
Institute of Computing Technology, \\
Chinese Academy of Sciences \\
Beijing, China
}
\addinstitution{
  University of Chinese Academy of \\
Sciences \\
Beijing, China
}

\runninghead{Zhao, Zeng, Shan}{Pose-Robust Point-of-Gaze Calibration}

\def\etal{\emph{et al}\bmvaOneDot}

\usepackage{enumitem}
\usepackage{multirow}
\usepackage{multicol}
\usepackage{booktabs}
\usepackage{wrapfig}
\usepackage{amssymb}
\usepackage{pifont}
\usepackage{makecell} 

\begin{document}

\maketitle

\begin{abstract}

Although appearance-based point-of-gaze (PoG) estimation has improved,  the estimators still struggle to generalize across individuals due to personal differences. Therefore, person-specific calibration is required for accurate PoG estimation. However, calibrated PoG estimators are often sensitive to head pose variations. 
To address this, we investigate the key factors influencing calibrated estimators and explore pose-robust calibration strategies. 
Specifically,  we first construct a benchmark, MobilePoG, which includes facial images from 32 individuals focusing on designated points under either fixed or continuously changing head poses. Using this benchmark, we systematically analyze how the diversity of calibration points and head poses influences estimation accuracy.
Our experiments show that introducing a wider range of head poses during calibration improves the estimator’s ability to handle pose variation. 
Building on this insight, we propose a dynamic calibration strategy 
in which users fixate on calibration points while moving their phones. 
This strategy naturally introduces head pose variation during a user-friendly and efficient calibration process, ultimately producing a better calibrated PoG estimator that is less sensitive to head pose variations than those using conventional calibration strategies.
Codes and datasets are available at our project page: https://mobile-pog.github.io.

\end{abstract}
\section{Introduction}
\label{sec:intro}

Gaze estimation has attracted growing interest in computer vision due to its ability to reveal human attention, intention, and cognitive processes.
Accurate point-of-gaze (PoG) prediction enables applications in medical diagnosis~\cite{Medical_1, Medical_2, Medical_3}, human-computer interaction~\cite{HCI_1, HCI_2, HCI_3, HCI_4}, VR/AR~\cite{VR_1, VR_2}, assisted driving~\cite{Dirving_1}, and so on.
Powered by deep learning methods, appearance-based PoG estimation predicts 2D gaze coordinates on a designated screen from a single RGB image captured by a front-facing camera, without requiring additional hardware such as LED lights, infrared sensors, or wearable eye trackers.

\begin{figure}[t]
    \centering
    \includegraphics[width=1\linewidth]{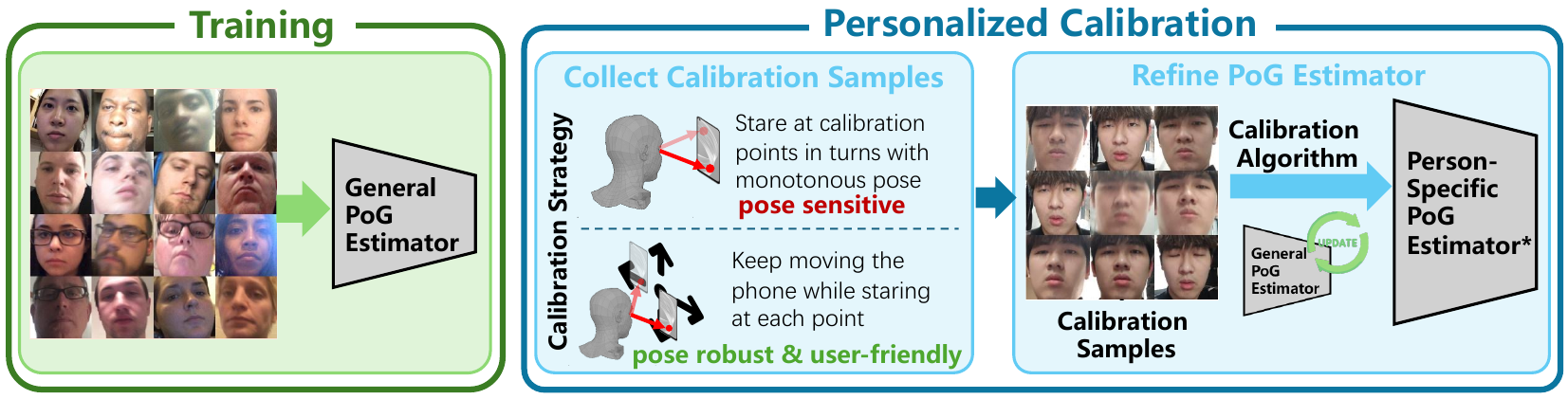}
    \caption{Pipeline of personalized calibration for PoG estimation.}
    \label{fig:teaser}
\end{figure}

Although appearance-based PoG estimation has advanced significantly with the development of large-scale datasets and data-driven learning methods, existing models still struggle to generalize across individuals due to challenging personal differences, such as variations in the anatomical structure of eyes~\cite{guestrin2006general, Differential}, that are often not discernible from RGB images alone.
To customize the PoG estimator for individual users, an effective and widely used approach is to perform personalized calibration before deployment~\cite{Differential, GazeCapture, Bias, Redirection, FAZE, Prompt}.  
As shown in Figure~\ref{fig:teaser}, rather than directly deploying a general PoG estimator trained on diverse data, personalized calibration typically begins by collecting a small set of user-specific samples, where the user is instructed to gaze at a few designated points (e.g., 5 or 9) while corresponding facial images are recorded.
These calibration samples are then used to adapt the general model to the individual, resulting in an improved personalized estimator.

However, real-world applications show that the performance of calibrated PoG estimators is highly sensitive to head pose variations.
When a user's head pose during inference differs from that during calibration, estimation accuracy often degrades significantly. 
This sensitivity is likely to stem from the static conventional calibration strategies, which require users to maintain a fixed head pose while gazing at predefined calibration points.
Despite its practical importance, the impact of calibration strategy on gaze estimation performance remains under-explored. 
A key limitation is that existing mobile PoG datasets lack head pose guidance during data collection, making it difficult to simulate diverse calibration scenarios under controlled conditions. 
Consequently, these datasets fall short in supporting rigorous evaluation of calibration strategies and their effectiveness.

To develop a calibrated PoG estimator that remains accurate across varying head poses, we investigate key calibration factors (i.e., the diversity of calibration points and head poses) and propose a pose-robust, user-friendly calibration strategy. 
We introduce MobilePoG, a new phone-based PoG dataset that captures facial images of users when they are fixating on predefined points under either fixed or continuously changing head poses. MobilePoG enables the simulation of diverse calibration scenarios, serving as a benchmark for evaluating calibration strategies and analyzing influential factors systematically. 
Experimental analysis on MobilePoG reveals that increasing head pose diversity in calibration samples significantly enhances estimator robustness, whereas simply adding more calibration points offers limited improvement. 
Based on this observation, we propose a dynamic calibration strategy, where users keep moving the phone while fixating on each calibration point, instead of maintaining a static pose. 
This simple yet effective strategy introduces natural head pose variation and enhances the robustness of the calibrated model across different head poses, regardless of the underlying PoG estimators or calibration algorithms. 
The contributions of this work are summarized as follows:
\begin{itemize}[leftmargin=1.5em]
  \item We construct MobilePoG, a mobile PoG dataset that enables the simulation of diverse calibration scenarios. It serves as a benchmark for evaluating calibration strategies and enables systematic analysis of key influencing factors. 
  \item We systematically analyze the impact of calibration points and head poses, and find that pose diversity in calibration samples plays a crucial role in enhancing model robustness.
  \item We propose a dynamic calibration strategy that naturally introduces head pose variation through a user-friendly and efficient process. Experiments on MobilePoG show that the proposed strategy effectively reduces the calibrated estimator's sensitivity to head pose.
\end{itemize}

\section{Related Work}
\label{sec:related work}

In this section, we review prior works on appearance-based gaze estimation methods, personalized gaze calibration, and point-of-gaze (PoG) datasets relevant to our approach.

\textbf{Appearance-based Gaze Estimation Methods:} Appearance-based methods aim to directly learn a mapping from input images to either a 3D gaze direction vector or a 2D point-of-gaze (PoG) on a screen. 
For 3D gaze estimation,
Zhang \etal~\cite{MPIIGaze} introduced a convolutional neural network to extract gaze direction from eye images and head pose, significantly enhancing prediction accuracy. 
Since then, numerous studies~\cite{RT-GENE, PoseGaze, EyeAsymmetry, CrossEncoder, cai} have further advanced appearance-based 3D gaze estimation.
The estimated 3D gaze vector can be projected to 2D screen coordinates if the geometric relationship between the camera and screen is known. 
For direct estimation of PoG, Krafka \etal~\cite{GazeCapture} and Zhang \etal~\cite{MPIIFaceGaze} improved accuracy by incorporating spatial attention and demonstrated the advantages of using full-face images to capture head pose information. To mitigate overfitting, He \etal~\cite{SAGE} simplified the architecture by using only eye images and eye corner landmarks, which also encode head pose cues. Guo \etal~\cite{TAT} introduced a knowledge distillation training scheme, while Bao \etal~\cite{AFFNet} and Cheng \etal~\cite{GazeTransformer} enhanced feature fusion through facial guidance and Transformer-based attention, respectively. Balim \etal~\cite{EFE} proposed an end-to-end framework that predicts gaze direction and origin separately before computing PoG.
To further enhance the generalization of gaze estimators, many studies have explored unsupervised domain adaptation~\cite{RotationConsistent, Contrastive, Prompt, UnReGA} and domain generalization~\cite{PureGaze, GazeConsistentFeature, PCF}.

In spite of the improvements brought by the aforementioned studies to general gaze estimators, their generalization capability is still constrained by inter-individual differences. This paper focuses on personalized calibration for point-of-gaze estimation, which holds significant value for practical applications.


\textbf{Personalized Gaze Calibration:} Personalized calibration tailors the gaze estimation model to an individual user by adjusting model parameters or outputs based on user-specific data to improve accuracy. 
Many studies focused on the calibration algorithms for adapting gaze estimators given user-specific calibration samples~\cite{Differential, GazeCapture, Bias, Redirection, FAZE, Prompt, SAGE}. 
These algorithms can be broadly categorized into three types: feature extraction-based methods, tuning-based methods, and anchor-based methods. 
Feature extraction-based methods utilize pre-trained models to extract features from calibration samples and then perform calibration using techniques such as support vector regression (SVR)~\cite{GazeCapture}, linear probe~\cite{Differential}, or bias correction~\cite{ Bias}, without updating the model weights.
In contrast, tuning-based methods~\cite{FAZE, Prompt} fine-tune a subset of the model parameters under the supervision of calibration samples. 
 Anchor-based methods~\cite{SAGE} treat calibration samples as reference anchors, enabling the model to learn predictions by incorporating the features and labels of anchors.

However, the strategy for collecting user-specific data has received comparatively less attention. 
Conventional strategies typically gather calibration samples while the user fixates on predefined calibration points with minimal variation in head pose. 
Some studies~\cite{GazeCapture, SAGE, Bias} have examined the influence of the number of calibration points, but often overlook head pose variation, which plays a critical role in determining the performance of calibration. 
In this work, we conduct a systematic analysis of the factors affecting calibration robustness and empirically demonstrate that head pose diversity during calibration is a key determinant for enhancing the generalization capability of point-of-gaze estimation.




\textbf{Point-of-Gaze Dataset:}
Many datasets~\cite{MPIIGaze, EyeDiap, UTMultiview, MPIIFaceGaze, GazeCapture, EVE, Gaze360, ETH-XGaze} have been built for the development of gaze estimation methods.
Some of them are annotated with 2D PoGs~\cite{UTMultiview, EyeDiap, TabletGaze, MPIIGaze, MPIIFaceGaze, GazeCapture, EVE} and have various subjects, PoGs, and head poses.
Although existing datasets have significantly advanced PoG estimation research, they are limited in supporting systematic studies of personalized PoG calibration. A key limitation is the lack of control over critical calibration conditions during data collection leads to difficulty in simulating real-world calibration. To enable rigorous analysis of calibration strategies, a dataset should provide explicit guidance for both gaze target locations and head pose configurations during the data acquisition process. Such control allows for well-designed experiments where the effects of individual factors (PoG and head pose) can be independently varied and studied. 
Take a few examples, MPIIFaceGaze~\cite{MPIIGaze, MPIIFaceGaze} collected real-world data using laptops in daily-life scenarios, where participants just adopted natural head poses as if they were using a computer in their usual manner. GazeCapture~\cite{GazeCapture}, the largest dataset with PoG on mobile phones and tablets, simply asks participants to switch device orientation and randomly change head pose. Without precise control over head pose configurations, they are unsuitable for studying gaze calibration.
TabletGaze~\cite{TabletGaze} collects PoGs on mobile devices covering predefined four static poses but the limited pose diversity and high label noise make it difficult to draw reliable and generalizable conclusions for calibration research.
Distinct from the previous datasets, we propose a benchmark specifically designed for personalized calibration in point-of-gaze estimation, which provides explicit guidance on head pose and point-of-gaze during data collection.





\section{MobilePoG Dataset}
\label{sec:dataset}
Since existing PoG datasets were not designed to investigate calibration factors and lack the necessary conditions for such analysis, we construct a new benchmark, MobilePoG, which contains facial images of individuals when they are fixating on designated points under either fixed or continuously changing head poses.

\begin{figure}[t]
    \centering
    \includegraphics[width=1.0\linewidth]{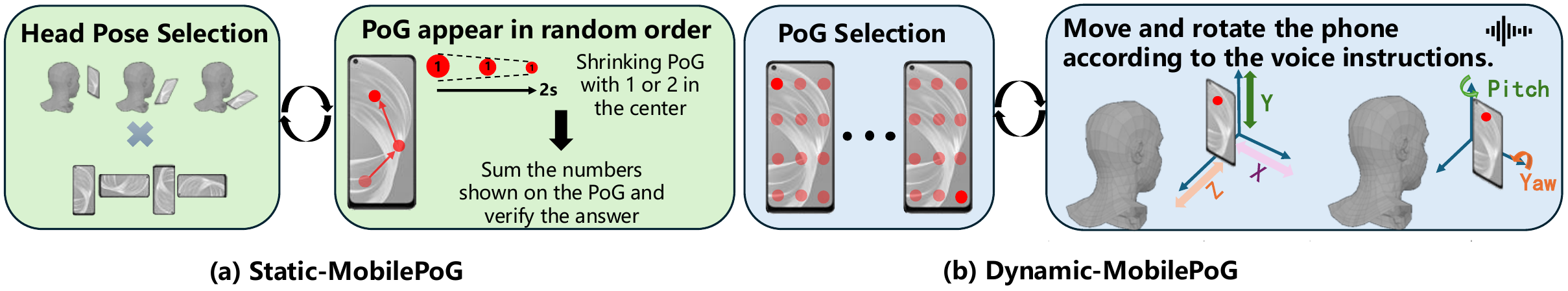}
    \caption{Dataset collection procedure of MobilePoG. 
    (a)Static-MobilePoG: Subjects fixate on the point with fixed poses.
    (b)Dynamic-MobilePoG: Subjects fixate on the point while moving their phones, resulting in continuously changing poses. }
    \label{fig:dataset_pipeline}
\end{figure}

\subsection{Dataset Collection}
We recruited 32 college students and recorded their facial images when they were instructed to fixate on designated points displayed on the phones. Two Android applications were developed to guide the collection procedure and form two subset of data: Static-MobilePoG with various fixed head poses, and Dynamic-MobilePoG with continuously changing poses.

\begin{figure}[t]
    \centering
    \includegraphics[width=0.8\linewidth]{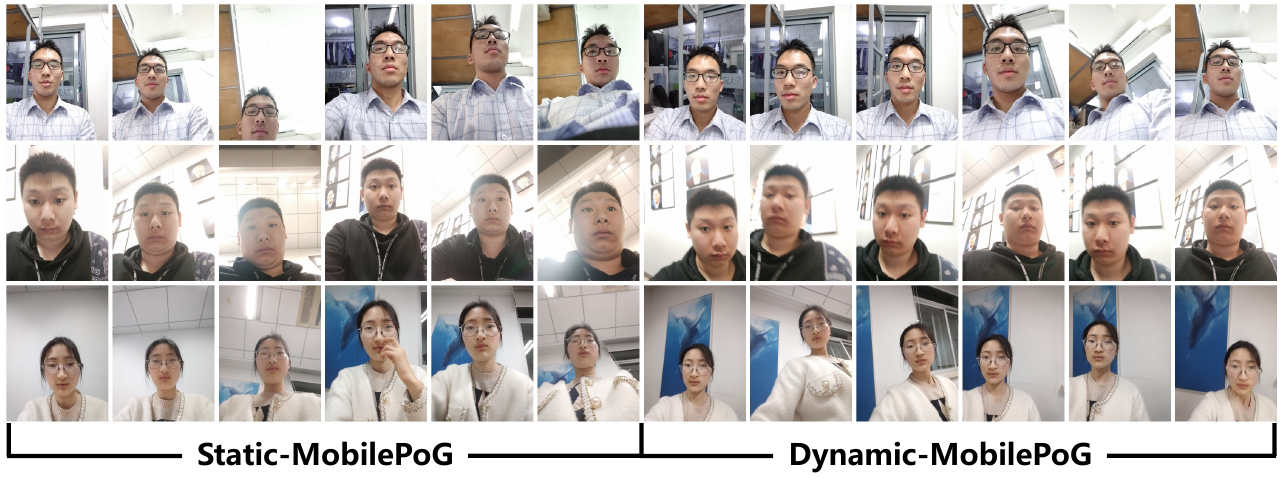}
    \caption{Examples from MobilePoG. 
    Each subject exhibits diverse head pose variations.}
    \label{fig:dataset_example}   
\end{figure}

\textbf{Static-MobilePoG:} As shown in Figure~\ref{fig:dataset_pipeline} (a), each participant is instructed to adopt one of three user-to-phone relative positions and to hold the phone in one of four orientations, yielding 12 head pose configurations commonly encountered during everyday smartphone use. For each pose, 55 predefined points appeared in random order. Participants are asked to fixate on each point while maintaining the same pose. 
Each point exists for 2 seconds and gradually shrinks in size to help guide accurate fixation.
Meanwhile, a digital number (either 1 or 2) was displayed at the center of each point, and participants were required to mentally accumulate the sum throughout the sequence.
The data was considered valid only if the participant reported the correct total after viewing all 55 points.
The left six columns in Figure~\ref{fig:dataset_example} show image samples with different poses in Static-MobilePoG.


\textbf{Dynamic-MobilePoG:} 
As shown in Figure~\ref{fig:dataset_pipeline} (b), for each of the predefined points, participants are instructed to fixate on the target point for a period of time. During this fixation period, they are guided by voice instructions to smoothly move and rotate the phone. These instructions induce head pose variations across five degrees of freedom: left-right ($x$), up-down ($y$), forward-backward ($z$), pitch, and yaw. Consequently, the collected images capture a comprehensive range of poses encountered during everyday mobile phone usage.
The right six columns in Figure~\ref{fig:dataset_example} show image samples with different poses in Dynamic-MobilePoG.


\subsection{Dataset Characteristics}

We recorded facial videos from 32 participants during both the Static-MobilePoG and Dynamic-MobilePoG procedures. For each point in Static-MobilePoG, we retained 30 frames from the middle one-second interval of the two-second duration. In Dynamic-MobilePoG, all frames for each point were retained. Facial landmarks were then detected using MediaPipe\cite{MediaPipe}. Frames with undetectable or blurry faces were excluded to ensure data quality. As a result, the MobilePoG dataset contains approximately 606,000 frames in the Static-MobilePoG subset and around 1.71 million frames in the Dynamic-MobilePoG subset.

\begin{table}[ht]
    \centering
    \renewcommand\tabcolsep{3.5pt}
    \begin{tabular}{l|ccccc}
      \toprule
      Dataset & \# Subject & \# PoG & \# Image & \# Pose & Pose Controllable\\
      \midrule
      TabletGaze~\cite{TabletGaze} & 51 & 35 & 1,785 & 4 Static & \ding{51} \\
      GazeCapture~\cite{GazeCapture} & 1468 & Random & 2,445,504 & Random & \ding{55} \\
      \midrule
      Static-MobilePoG & 32 & 55 & 606,301 & 12 Static & \ding{51} \\
      Dynamic-MobilePoG & 32 & 12 & 1,707,910 & Continuous & \ding{51} \\
      \bottomrule 
    \end{tabular}
    \caption{Comparison of existing mobile PoG datasets.}
    \label{table:dataset_comparison}
\end{table}

Table \ref{table:dataset_comparison} presents a statistical comparison of existing mobile PoG datasets. 
TabletGaze~\cite{TabletGaze} recorded images under only four poses for each subject.
GazeCapture\cite{GazeCapture} includes a large number of participants, contributing substantial diversity to the dataset.
However, each subject in GazeCapture looked at random points while maintaining self-selected, comfortable poses. This randomness makes it difficult to isolate and study the effects of individual calibration factors, as multiple variables change simultaneously.
In contrast, MobilePoG provides controlled yet diverse pose variations, particularly in the Dynamic-MobilePoG subset, where phone movement results in a nearly continuous range of head poses. The dataset includes images of participants fixating on the same point from different poses, as well as images captured under the same pose while fixating on different points. This design facilitates systematic analysis of how individual factors affect calibration.
\section{Factor Analysis of Personalized PoG Calibration}
\label{sec:smpg}

\subsection{Personalized PoG Calibration Pipeline}
Figure~\ref{fig:teaser} illustrates the pipeline of personalized point-of-gaze (PoG) calibration. In the training stage, a general PoG estimator is first trained on data collected from multiple subjects. Rather than using this general model directly for inference, we adapt it to individual users through personalization calibration that typically involves two steps. First, a small set of calibration samples is collected based on a chosen calibration strategy. For example, the conventional static strategy asks the user to fixate on a few designated points (e.g., 5 or 9) while corresponding facial images are captured. Second, these samples are used to refine the general estimator via a calibration algorithm, resulting in a person-specific PoG model that is then deployed for that individual.
\subsection{Experimental Setup}
To understand how different factors influence the performance of personalized point-of-gaze (PoG) calibration, we conducted a series of controlled experiments using the Static-MobilePoG. We focused on two key variables: the number of calibration points and the diversity of head poses during calibration.

\textbf{General PoG Estimator:} To ensure the generality of our analysis, we adopted two prevalent methods iTracker~\cite{GazeCapture} and AFFNet~\cite{AFFNet} as the general PoG estimator. The general PoG estimator was initially pre-trained on the GazeCapture~\cite{GazeCapture} dataset to endow it with stronger generalization ability and a solid foundational representation, and was subsequently trained on the training set of Static-MobilePoG.
In Static-MobilePoG, we randomly selected 5 subjects as a calibration set and the remaining 27 subjects as the training set. 

\textbf{Calibration Samples:} We investigated varied configurations of calibration points and head poses in the conventional static strategy. For each subject in the calibration set, the calibration samples are selected as the ones staring at $N$ predefined points ($N=1,5,9,13$) with $P$ poses ($P=1,2,3,4$). The position of points is illustrated in the supplemental materials.

\textbf{Calibration Algorithm:} Different calibration algorithms are adopted for calibration-algorithm-agnostic findings. The algorithms include: (1)\textit{SVR} replaces the model's MLP head to regress the PoG. (2)\textit{Linear Probe} trains a linear layer on top of the model. (3)\textit{Finetune MLP} fine-tunes the MLP regression head. (4)\textit{Full Finetune} fine-tunes the whole estimator. For the detailed implementation of algorithms, please refer to the supplementary materials.



\subsection{Monotonous Pose Calibration Limits Generalization}
To investigate the bottlenecks in real-world calibration scenarios, we conducted experiments using a monotonous head pose with multiple calibration points, simulating the conventional static calibration strategy. Table~\ref{table:single_pose} reports the performance of calibrated estimators based on samples collected under a single head pose.

%

As shown in Table~\ref{table:single_pose}, increasing the number of calibrated points under a fixed pose leads to a substantial reduction in the same pose estimation error, with performance approaching saturation at 9 calibration points. For instance, when calibrated with 9 points using the Full Finetune strategy, AFFNet achieves a 44.1\% error reduction, reaching 1.42 cm.

However, despite the improved accuracy under the same head pose, the calibrated estimators generalize poorly to test samples with different head poses, resulting in high average errors. This issue is particularly pronounced for SVR and Linear Probe, both of which train a regression head from scratch. In these cases, single-head-pose calibration samples lead to severe overfitting. Consequently, the minimum average errors for AFFNet and iTracker calibrated using SVR and Linear Probe reach 7.93 cm / 3.19 cm and 7.92 cm / 4.00 cm, respectively, which are even worse than those without any calibration. While Finetune MLP and Full Finetune exhibit slightly better robustness, their performance gains are still limited due to the narrow distribution of calibration poses. For example, calibrating AFFNet with 13 points using Full Finetune only reduces the error from 2.54 cm to 2.30 cm.

These results clearly indicate that calibration based on monotonous head pose samples fails to generalize to varying head poses. Thus, incorporating pose diversity in the calibration data is essential for achieving robust and reliable PoG estimation in real-world scenarios.


\subsection{Diverse Pose Calibration Boosts Generalization}
\begin{figure}[t]
    \centering
    \includegraphics[width=1.0\linewidth]{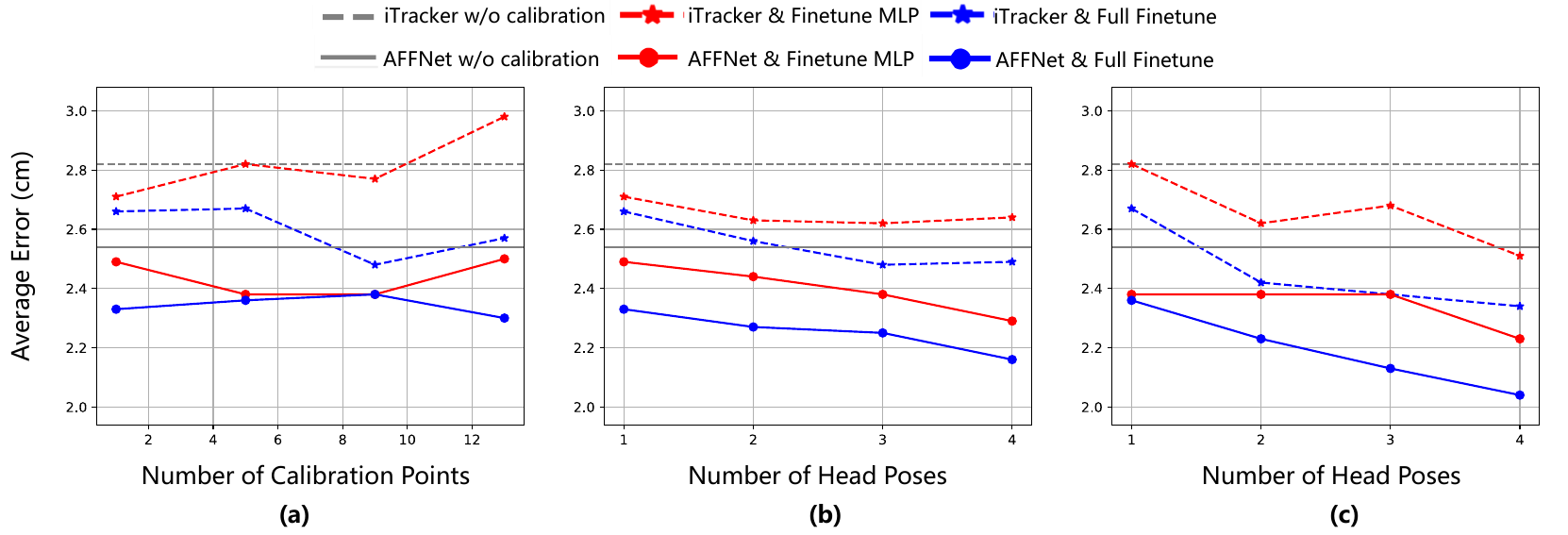}
    \caption{Comparison of increasing calibration points and head poses in calibration samples. (a) shows increasing points with a single head pose. (b) shows increasing head poses with one point. (c) shows increasing head poses with five points.}
    \label{fig:pog_vs_pose}
\end{figure}

To further investigate the key factors influencing the robustness of calibrated estimators, we compared performance variations under two conditions: increasing the number of calibration points (PoGs) and increasing the diversity of calibration head poses. Figure~\ref{fig:pog_vs_pose} illustrates how the average error of the calibrated estimator changes with increasing numbers of calibration points and head poses in the calibration samples.

As shown in Figure~\ref{fig:pog_vs_pose} (a), when the head pose in calibration samples is fixed, increasing the number of PoGs yields only marginal performance gains. In some cases, adding too many PoGs even leads to overfitting. For instance, iTracker calibrated via Finetune MLP with thirteen PoGs performs worse than the uncalibrated baseline.

In contrast, Figure~\ref{fig:pog_vs_pose} (b) and (c) reveal that increasing head pose diversity in calibration samples significantly improves the robustness of calibrated estimators. Notably, even with only a single point, calibration under multiple (3 or 4) head poses consistently outperforms the case with multiple (9 or 13) points under a single head pose, as shown in Figure~\ref{fig:pog_vs_pose} (b).

These findings strongly suggest that head pose diversity is a critical factor in personalized calibration. Incorporating varied head poses into calibration samples substantially enhances the generalization ability of calibrated estimators and should be prioritized when designing practical calibration strategies.



\begin{table}[t]
\centering
\renewcommand\tabcolsep{3.5pt}
\scriptsize
\begin{tabular}{l|c|ccc|ccc|ccc|ccc}
\toprule
\multirow{2}{*}{\textbf{Model}} & \multirow{2}{*}{\textbf{ \makecell{Calibration \\ Algorithm}}} & \multicolumn{3}{c|}{\textbf{1 calibration point}} & \multicolumn{3}{c|}{\textbf{5 calibration points}} & \multicolumn{3}{c|}{\textbf{9 calibration points}} & \multicolumn{3}{c}{\textbf{13 calibration points}} \\
& & Same & Diff. & Avg. & Same & Diff. & Avg. & Same & Diff. & Avg. & Same & Diff. & Avg.\\
\midrule

\multirow{5}{*}{\textbf{iTracker}} 
& w/o calibration & \multicolumn{12}{c}{\cellcolor{gray!20} 2.82} \\
& SVR & 4.17 & 11.17 & 10.60 & 2.30 & 8.93 & 8.42 & 2.21 & 8.47 & 8.02 & 2.27 & 8.31 & 7.92 \\
& Linear Probe & 3.82 & 7.08 & 6.81 & 2.40 & 4.13 & 4.00 & 2.17 & 5.30 & 5.07 & 2.29 & 4.58 & 4.43 \\
& Finetune MLP & 2.35 & 2.74 & 2.71 & 2.10 & 2.88 & 2.82 & 2.00 & 2.83 & 2.77 & 2.02 & 3.05 & 2.98 \\
& Full Finetune & 2.10 & 2.71 & 2.66 & 2.07 & 2.72 & 2.67 & 1.97 & 2.63 & 2.58 & 1.40 & 2.65 & 2.57\\
\midrule

\multirow{5}{*}{\textbf{AFFNet}}
& w/o calibration & \multicolumn{12}{c}{\cellcolor{gray!20} 2.54} \\
& SVR & 4.18 & 11.17 & 10.60 & 2.18 & 8.40 & 7.93 & 1.69 & 8.91 & 8.40 & 2.03 & 8.60 & 8.17 \\
& Linear Probe & 4.18 & 6.84 & 6.62 & 2.02 & 3.29 & 3.19 & 1.97 & 3.69 & 3.57  & 2.05 & 3.55 & 3.45 \\
& Finetune MLP & 2.06 & 2.53 & 2.49 & 2.04 & 2.41 & 2.38 & 1.52 & 2.45 & 2.38 & 1.48 & 2.57 & 2.50 \\
& Full Finetune & 1.94 & 2.37 & 2.33 & 1.99 & 2.39 & 2.36 & 1.42 & 2.45 & 2.38 & 1.40 & 2.37 & 2.30 \\
\bottomrule

\end{tabular}
\caption{Results of a single calibration head pose and different numbers of calibration points on Static-MobilePoG. "Same" and "Diff." represent the test errors of the calibrated model on test samples with the same and different head poses as the calibration head pose, respectively. The metric is the Euclidean distance (cm).}
\label{table:single_pose}
\end{table}


\section{Pose-Robust Calibration Strategy}
\label{sec:dmpg}

\subsection{Dynamic Calibration Strategy Design}
As shown in Section~\ref{sec:smpg}, head pose diversity in calibration samples greatly impacts the performance of person-specific estimators. However, increasing this diversity complicates data collection and demands more user cooperation. Thus, finding a user-friendly calibration strategy is crucial for practical deployment on mobile devices.

As shown in Figure~\ref{fig:dataset_pipeline} (b), we propose a new dynamic calibration strategy. For each calibration PoG, the user just needs to rotate and translate the mobile phone within a range while keeping gaze at the PoG, which not only allows for the collection of calibration samples with sufficiently diverse head poses, but is also user-friendly and easy to operate on mobile devices. As mentioned in Section~\ref{sec:dataset}, the collection procedure of Dynamic-MobilePoG could efficiently simulate the proposed dynamic calibration strategy as the dataset contains continuously varying head poses. We will conduct experiments on Dynamic-MobilePoG to demonstrate that our proposed dynamic strategy outperforms the conventional static strategy.

\subsection{Experimental Evaluation}
\subsubsection{Experiment Settings}
We evaluated the proposed strategy on Dynamic-MobilePoG.
Five subjects were randomly selected for the calibration set, while the remaining 27 subjects were used for training.
Similar to settings in Section~\ref {sec:smpg}, we trained the general PoG estimator by first pretraining either iTracker~\cite{GazeCapture} or AFFNet~\cite{AFFNet} on the GazeCapture dataset, and subsequently trained them on the Dynamic-MobilePoG training set.

To compare the conventional static strategy and our proposed dynamic strategy, we conducted experiments on Dynamic-MobilePoG by selecting calibration samples with different settings to simulate the two strategies. 
For the \textbf{static strategy}, we selected consecutive frames from the sequence as calibration data and used the remaining data for testing.
For the \textbf{dynamic strategy},  we sampled from the frame sequence at a fixed step to ensure that selected frames have diverse head poses and evaluated the calibrated model on the rest of the data for each person.
We conducted experiments with 1, 2, 4, and 6 calibration points for both strategies, and sampled 30 frames for each calibration point. The position of points is illustrated in the supplemental materials.

Finally, to ensure the generalizability of the experimental conclusions, we also employed a range of calibration algorithms (i.e., SVR, Linear Probe, Finetune MLP, and Full Finetune) to construct personalized PoG estimators.

\subsubsection{Experimental Results}

\begin{table}[t]
\scriptsize
\centering
\renewcommand\tabcolsep{1pt}
\begin{tabular}{l|c|cc|cc|cc|cc}
\toprule
\multirow{2}{*}{\textbf{Model}} & \multirow{2}{*}{\textbf{ \makecell{Calibration \\ Algorithm}}} & \multicolumn{2}{c|}{\textbf{1 calib. point}} & \multicolumn{2}{c|}{\textbf{2 calib. points}} & \multicolumn{2}{c|}{\textbf{4 calib. points}} & \multicolumn{2}{c}{\textbf{6 calib. points}}\\
& & static & dynamic & static & dynamic & static & dynamic & static & dynamic \\
\midrule

\multirow{5}{*}{\textbf{iTracker}} 
& w/o calibration & \multicolumn{8}{c}{\cellcolor{gray!20} 2.09} \\
& SVR & 4.18 & \textbf{4.18} & 3.35 & \textbf{3.11} & 2.26 & \textbf{1.89} & 2.17 & \textbf{1.76} \\
& Linear Probe & 2.86 & \textbf{2.83} & 2.62 & \textbf{2.58} & 2.03 & \textbf{1.89} & 1.97 & \textbf{1.77} \\
& Finetune MLP & 2.55 & \textbf{1.81} & 2.24 & \textbf{1.89} & 2.02 & \textbf{1.63} & 1.95 & \textbf{1.53} \\
& Full Finetune & 1.86 & \textbf{1.85} & 1.87 & \textbf{1.74} & 1.95 & \textbf{1.48} & 1.83 & \textbf{1.36} \\
\midrule

\multirow{5}{*}{\textbf{AFFNet}}
& w/o calibration & \multicolumn{8}{c}{\cellcolor{gray!20} 1.75} \\
& SVR & 4.18 & \textbf{4.18} & 3.19 & \textbf{2.9} & 1.60 & \textbf{1.39} & 1.60 & \textbf{1.35} \\
& Linear Probe & 2.77 & \textbf{2.67} & 2.35 & 2.41 & 1.69 & \textbf{1.66} & 1.58 & \textbf{1.51} \\
& Finetune MLP & 1.79 & \textbf{1.57} & 1.83 & \textbf{1.53} & 1.52 & \textbf{1.33} & 1.40 & \textbf{1.27} \\
& Full Finetune & 1.49 & \textbf{1.43} & 1.56 & \textbf{1.42} & 1.83 & \textbf{1.27} & 1.52 & \textbf{1.10} \\
\bottomrule

\end{tabular}
\caption{Comparison of the static and the dynamic strategy on Dynamic-MobilePoG.The metric is the Euclidean distance (cm).}
\label{table:strategy}
\end{table}

\textbf{Calibration Performance of the Static and the Dynamic Strategy.} 
Table~\ref{table:strategy} reports the errors of the calibrated estimators using various configurations (i.e., two models, four calibration algorithms, and four different numbers of calibration points) under the static and the dynamic strategy. 
As can be seen, the dynamic strategy outperforms the static strategy in almost all configurations except for the situation of AFFNet calibrated by Linear Probe with two calibration points. We can also see that the dynamic strategy can even surpass the performance of the static strategy, which requires a large number of calibration PoGs, using only a small number of calibration PoGs. For example, the test error of iTracker calibrated by Finetune MLP with the dynamic strategy and only one PoG achieves 1.81 cm, while the test error of that with the static strategy and six PoGs is 1.95 cm. The above experimental results demonstrate that the dynamic strategy, which is capable of collecting calibration samples with rich pose diversity, significantly enhances the robustness of the model after calibration.

\begin{figure}[t]
    \centering
    \includegraphics[width=0.8\linewidth]{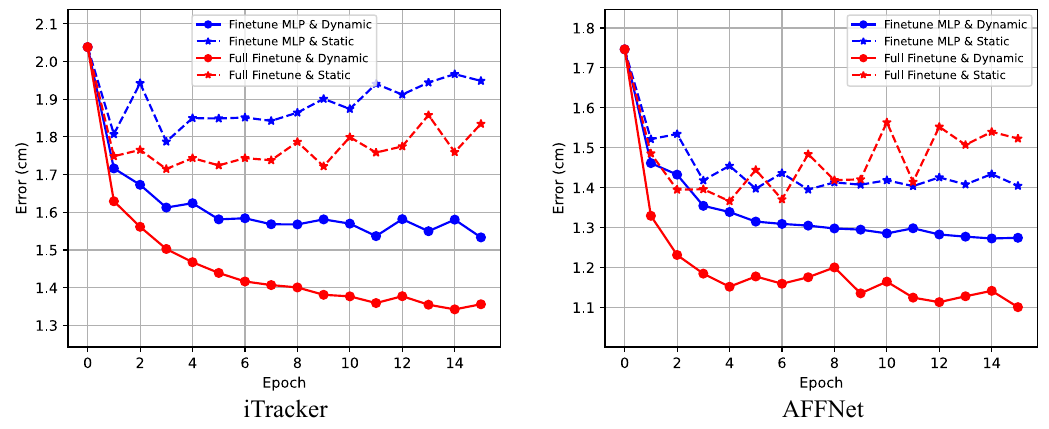}
    \caption{Calibration stability of the static and the dynamic strategy. All the models are calibrated with six calibration points.}
    \label{fig:stable}
     \vspace{-1 em}
\end{figure}

\textbf{Calibration Stability of the Static and the Dynamic Strategy.} To further demonstrate the efficiency of our proposed dynamic strategy, we evaluated the model performance stability of different estimators and calibration algorithms under two different calibration strategies. Figure~\ref{fig:stable} shows the performance of iTracker and AFFNet during calibration using Finetune MLP and Full Finetune across different epochs. We can observe that models calibrated using the dynamic strategy gradually converge to an optimal performance and remain stable, whereas models calibrated using the static strategy exhibit severe performance fluctuations. For models with unstable performance, achieving optimal results may require careful tuning of hyperparameters such as the learning rate or manually selecting a checkpoint from a specific epoch, which is not practical in real-world calibration applications. With our proposed dynamic strategy, the person-specific estimator exhibits not only excellent but also stable performance, enabling efficient personalized PoG calibration.

\vspace{-1 em}
\section{Conclusion}
\vspace{-0.5 em}
\label{conclusion}

In order to bridge the gap in research on personalized PoG calibration strategies and mitigate the sensitivity of the conventional static strategy to head pose variations, we first construct a new dataset, MobilePoG, which can faithfully simulate the calibration process and thus serves as a benchmark for personalized calibration. Our experiments on Static-MobilePoG demonstrate that the main bottleneck in personalized calibration lies in the diversity of head poses within the calibration samples. Finally, we propose a user-friendly dynamic calibration strategy and validate its effectiveness on Dynamic-MobilePoG, showing that it significantly improves the person-specific estimator’s performance and robustness to varying head poses.
Future work may focus on incorporating mechanisms for subsequent personalized calibration into the design of the general PoG estimator, particularly with respect to head pose variations, which could lead to more efficient calibration and improved robustness of the calibrated model.


\textbf{ACKNOWLEDGEMENT:} This work is supported by National Natural Science Foundation of China (No. 62176248, U2336213).

\clearpage

\bibliography{egbib}

\clearpage
\appendix

\section{More Discussion about Head Pose}
\label{sec:discussion}
In the main paper, we mention that the conventional static strategy is sensitive to head pose variation and demonstrate that the calibration samples' head pose diversity has a lot of impact on the person-specific PoG estimator. In this section, we will have more discussion about the head pose in PoG estimation and calibration.

\begin{figure}[h]
    \centering
    \includegraphics[width=0.9\linewidth]{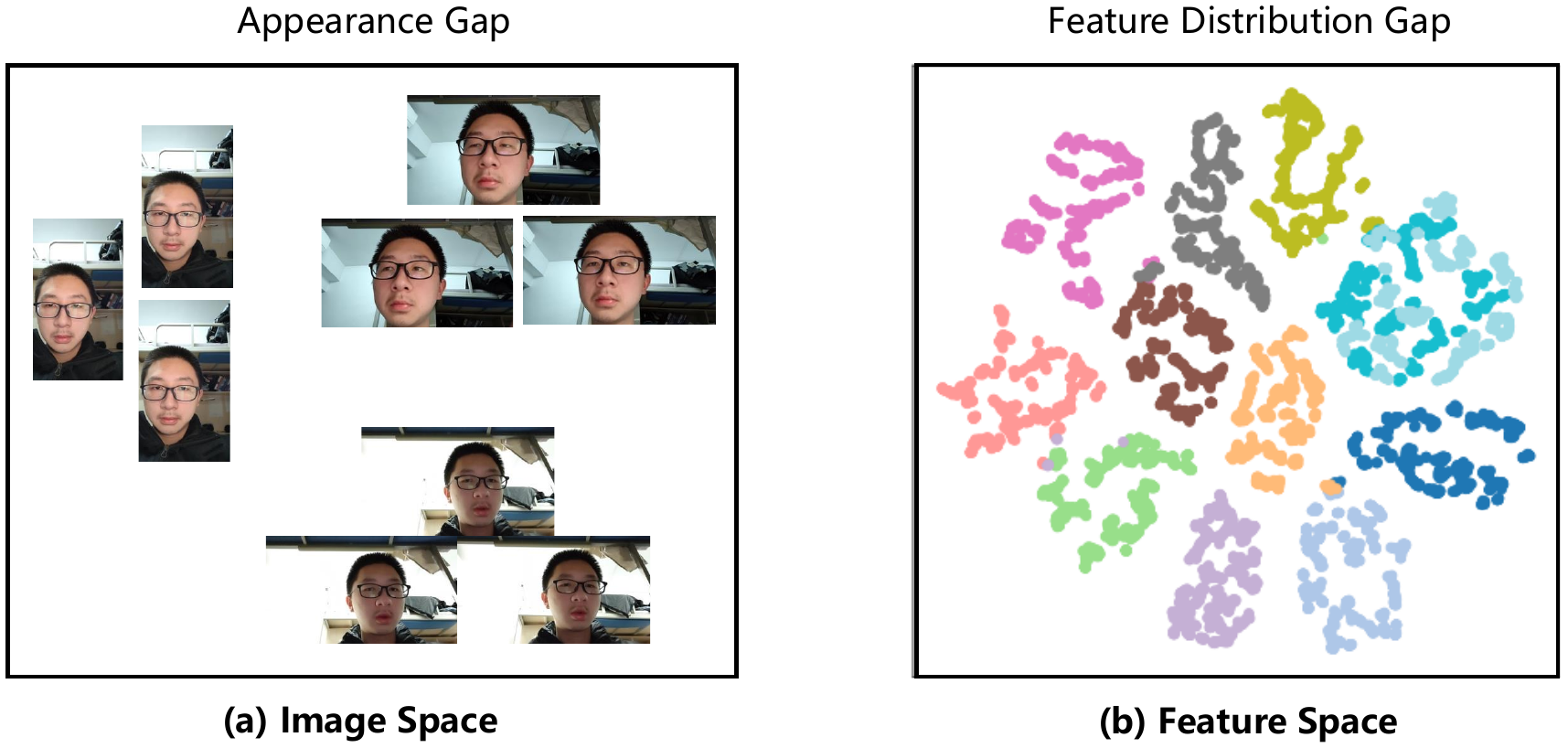}
    \caption{Gap between different head poses. (a) shows the appearance gap in the image space. (b) shows the distribution gap in the feature space with t-SNE.}
    \label{fig:gap}
\end{figure}

Figure~\ref{fig:gap} shows the gap between different head poses from Static-MobilePoG. Figure~\ref{fig:gap} (a) shows the appearance gap in the image space. We can observe that there is a significant gap between different head poses. When we use some calibration algorithms like Full Finetune, calibration samples with limited pose diversity can lead to model overfitting because of the appearance gap, making it less robust to pose variations. In Figure~\ref{fig:gap} (b), we perform t-SNE on one subject's feature extracted by a trained general estimator, where each color represents a head pose in Static-MobilePoG. We can see that the learned features exhibit clustering patterns based on head pose. In the personalized calibration stage, if we adopt some algorithms like SVR and Finetune MLP, they freeze the estimator's backbone and refine the regression head. This is equivalent to feeding the features into the model as input. Thus, such clustering patterns shown in Figure~\ref{fig:gap} (b) indicate that a single head pose will cause overfitting. Overall, head pose diversity is a principal factor driving personalized calibration performance.

\section{More MobilePoG's Visualization}
\begin{figure}[h]
    \centering
    \includegraphics[width=1\linewidth]{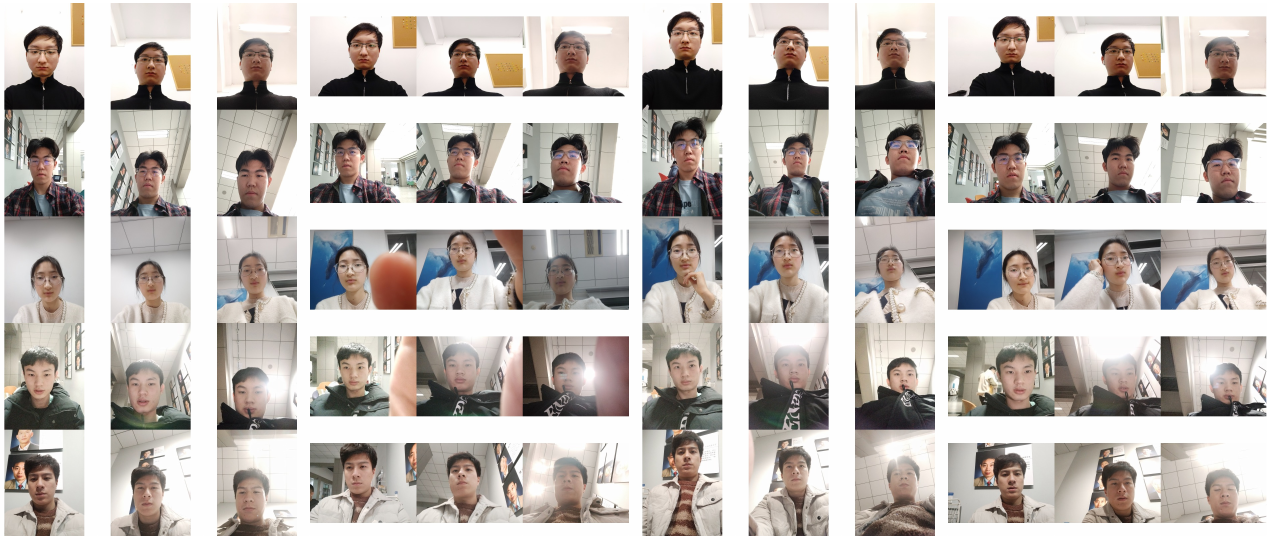}
    \caption{Visualization of Static-MobilePoG. 12 columns correspond to 12 types of head poses.}
    \label{fig:more_examples_static}
\end{figure}

\begin{figure}[h]
    \centering
    \includegraphics[width=1\linewidth]{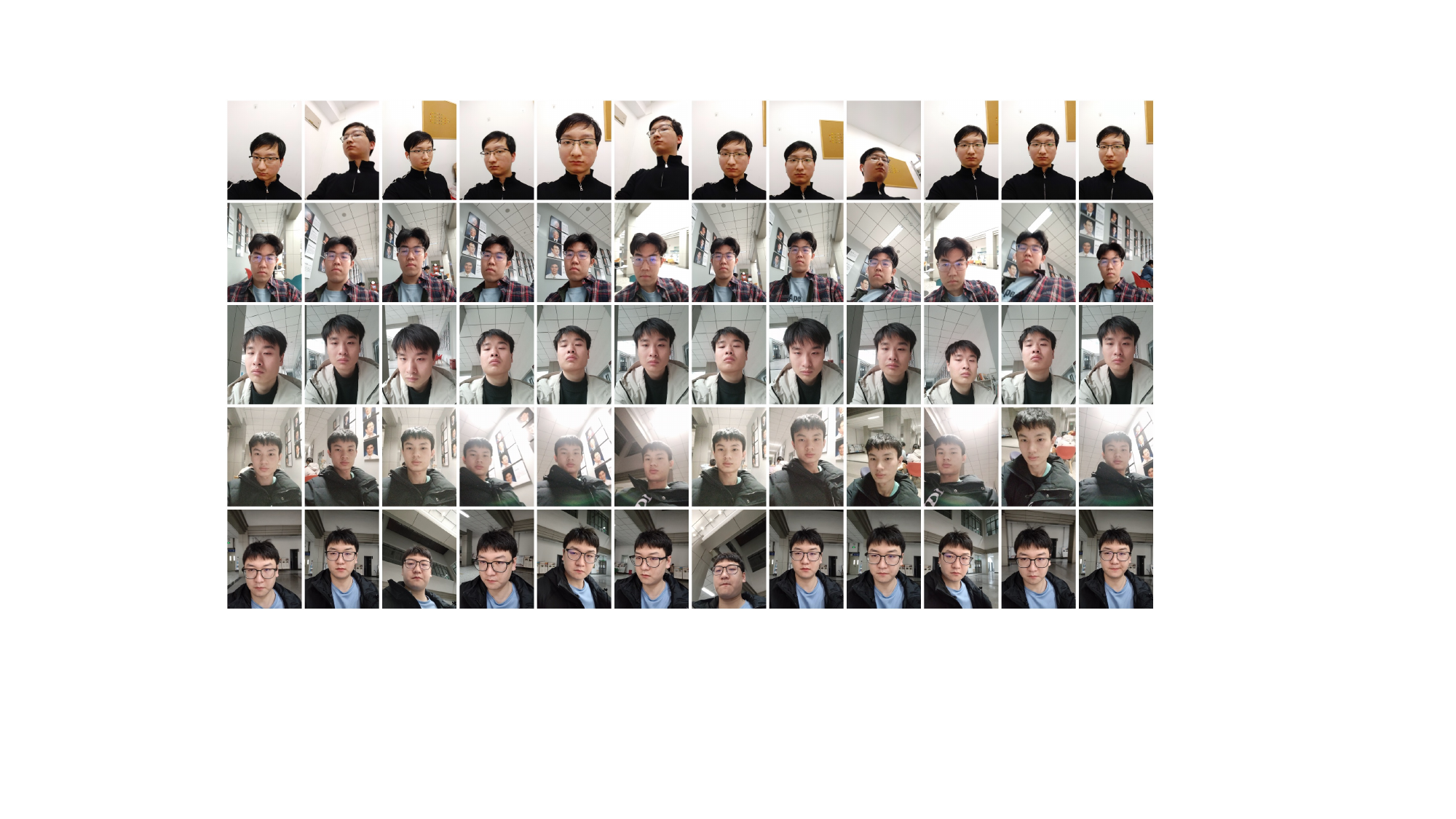}
    \caption{Visualization of Dynamic-MobilePoG. We can see the continuously varying head poses in each row.}
    \label{fig:more_examples_dynamic}
\end{figure}

In this section, we display more visualizations of the MobilePoG dataset. Figure~\ref{fig:more_examples_static} shows some examples from Static-MobilePoG, which contains 12 types of static head poses. Figure~\ref{fig:more_examples_dynamic} shows some examples from Dynamic-MobilePoG, which has continuously varying head pose in the frame sequences.
It can be observed that the MobilePoG dataset exhibits rich head pose diversity, making it well-suited for evaluating a model’s robustness to head pose variations under different experimental settings.

\section{More Experiment Details}

\subsection{Experiment Settings}
In this subsection, we provide more details about the settings of the main paper's experiments. 

\textbf{Estimator:} We selected \textit{iTracker}~\cite{GazeCapture} and \textit{AFFNet}~\cite{AFFNet} as our estimator since they are respectively the representative baseline and the state-of-the-art model in PoG estimation. They both consist of a CNN-based backbone and an MLP regression head. 

\begin{figure}[h]
    \centering
    \includegraphics[width=0.4\linewidth]{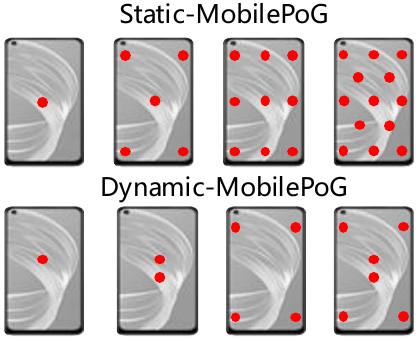}
    \caption{Locations of calibration points.}
    \label{fig:cal_pog}
\end{figure}

\textbf{Calibration PoG:} Figure~\ref{fig:cal_pog} shows the locations of calibration points we used in our experiments. The calibration point positions are distributed as uniformly as possible across the phone screen, aiming to achieve a more comprehensive coverage of the calibration space.

\subsection{Calibration algorithms:} 
In this subsection, we provide more details about the calibration algorithms used in our experiments.

\textbf{SVR} is a regression method that applies the principles of SVM to predict continuous values by fitting a function within a specified error margin. To apply SVR for PoG calibration, we replace the models' MLP head with an SVR to regress the PoG from the features extracted by the backbone following~\cite{GazeCapture}. 

\textbf{Linear Probe} makes predictions by training a simple linear layer on top while keeping the backbone frozen. Liu \etal~\cite{Differential} propose that the bias between the ground truth and the predictions can be estimated by a linear model. We extend this conclusion from 3D gaze direction to the 2D PoG, and during calibration, we add a linear layer on top of the model while freezing the rest of the parameters. 

\textbf{Finetune MLP} is to freeze the parameters of the feature extraction backbone during calibration and only learn the parameters of the MLP head. This approach balances model performance and training efficiency, which is a common algorithm in fine-tuning and transfer learning for vision models. We adopt the AdamW optimizer and train for 15 epochs on calibration samples. The learning rate on Static-MobilePoG is 1e-4, and the learning rate on Dynamic-MobilePoG is 5e-4.

\textbf{Full Finetune} trains all the model parameters on the new dataset. Although this involves a larger number of parameters, it can significantly improve the model’s transfer performance. We adopt the AdamW optimizer and train for 15 epochs on calibration samples. The learning rate on Static-MobilePoG is 1e-5, and the learning rate on Dynamic-MobilePoG is 5e-5.

There are still some other calibration algorithms like Prompt Tuning~\cite{Prompt} and Anchor Method~\cite{SAGE}. We don't select these algorithms because they need a more complex pipeline, which makes them difficult to flexibly adjust the calibration settings on a large scale. As discussed in Section~\ref{sec:discussion}, the inherent gap between different head poses still limits the performance of these algorithms if we adopt the conventional static strategy.

\subsection{All Experiment Results on Static-MobilePoG}
In this subsection, we provide all the experiment results on Static-MobilePoG, which are not shown in the main paper.

The following tables are directly related to Table 2 and Figure 4 in the main manuscript. They provide extended results and additional details to support the key findings. For a comprehensive analysis and discussion of the experiments, please refer to Section 4 of the main paper.

\begin{table}[ht]
\centering
\renewcommand\tabcolsep{3.5pt}
\scriptsize
\begin{tabular}{l|c|ccc|ccc|ccc|ccc}
\toprule
\multirow{2}{*}{\textbf{Model}} & \multirow{2}{*}{\textbf{ \makecell{Calibration \\ Algorithm}}} & \multicolumn{3}{c|}{\textbf{1 calibration point}} & \multicolumn{3}{c|}{\textbf{5 calibration points}} & \multicolumn{3}{c|}{\textbf{9 calibration points}} & \multicolumn{3}{c}{\textbf{13 calibration points}} \\
& & Same & Diff. & Avg. & Same & Diff. & Avg. & Same & Diff. & Avg. & Same & Diff. & Avg.\\
\midrule

\multirow{5}{*}{\textbf{iTracker}} 
& w/o calibration & \multicolumn{12}{c}{\cellcolor{gray!20} 2.82} \\
& SVR & 4.17 & 11.17 & 10.60 & 2.30 & 8.93 & 8.42 & 2.21 & 8.47 & 8.02 & 2.27 & 8.31 & 7.92 \\
& Linear Probe & 3.82 & 7.08 & 6.81 & 2.40 & 4.13 & 4.00 & 2.17 & 5.30 & 5.07 & 2.29 & 4.58 & 4.43 \\
& Finetune MLP & 2.35 & 2.74 & 2.71 & 2.10 & 2.88 & 2.82 & 2.00 & 2.83 & 2.77 & 2.02 & 3.05 & 2.98 \\
& Full Finetune & 2.10 & 2.71 & 2.66 & 2.07 & 2.72 & 2.67 & 1.97 & 2.63 & 2.58 & 1.40 & 2.65 & 2.57\\
\midrule

\multirow{5}{*}{\textbf{AFFNet}}
& w/o calibration & \multicolumn{12}{c}{\cellcolor{gray!20} 2.54} \\
& SVR & 4.18 & 11.17 & 10.60 & 2.18 & 8.40 & 7.93 & 1.69 & 8.91 & 8.40 & 2.03 & 8.60 & 8.17 \\
& Linear Probe & 4.18 & 6.84 & 6.62 & 2.02 & 3.29 & 3.19 & 1.97 & 3.69 & 3.57  & 2.05 & 3.55 & 3.45 \\
& Finetune MLP & 2.06 & 2.53 & 2.49 & 2.04 & 2.41 & 2.38 & 1.52 & 2.45 & 2.38 & 1.48 & 2.57 & 2.50 \\
& Full Finetune & 1.94 & 2.37 & 2.33 & 1.99 & 2.39 & 2.36 & 1.42 & 2.45 & 2.38 & 1.40 & 2.37 & 2.30 \\
\bottomrule

\end{tabular}
\caption{Results of a single calibration head pose and different numbers of calibration points on Static-MobilePoG. "Same" and "Diff." represent the test errors of the calibrated model on test samples with the same and different head poses as the calibration head pose, respectively. The metric is the Euclidean distance (cm).}
\label{table:single_pose}
\end{table}

\begin{table}[ht]
\centering
\renewcommand\tabcolsep{3.5pt}
\scriptsize
\begin{tabular}{l|c|ccc|ccc|ccc|ccc}
\toprule
\multirow{2}{*}{\textbf{Model}} & \multirow{2}{*}{\textbf{ \makecell{Calibration \\ Algorithm}}} & \multicolumn{3}{c|}{\textbf{1 calibration point}} & \multicolumn{3}{c|}{\textbf{5 calibration points}} & \multicolumn{3}{c|}{\textbf{9 calibration points}} & \multicolumn{3}{c}{\textbf{13 calibration points}} \\
& & Same & Diff. & Avg. & Same & Diff. & Avg. & Same & Diff. & Avg. & Same & Diff. & Avg.\\
\midrule

\multirow{5}{*}{\textbf{iTracker}} 
& w/o calibration & \multicolumn{12}{c}{\cellcolor{gray!20} 2.82} \\
& SVR & 4.15 & 8.70 & 7.95 & 2.31 & 5.75 & 5.22 & 2.13 & 4.87 & 4.48 & 2.01 & 5.34 & 4.90 \\
& Linear Probe & 3.25 & 4.48 & 4.28 & 2.24 & 3.13 & 2.99 & 2.54 & 3.29 & 3.18 & 2.47 & 3.95 & 3.75 \\
& Finetune MLP & 2.53 & 2.65 & 2.63 & 2.24 & 2.69 & 2.62 & 1.84 & 2.75 & 2.62 & 1.76 & 2.64 & 2.52 \\
& Full Finetune & 2.46 & 2.58 & 2.56 & 1.70 & 2.56 & 2.42 & 1.69 & 2.53 & 2.41 & 1.76 & 2.45 & 2.36 \\
\midrule

\multirow{5}{*}{\textbf{AFFNet}}
& w/o calibration & \multicolumn{12}{c}{\cellcolor{gray!20} 2.54} \\
& SVR & 4.03 & 7.82 & 7.19 & 1.82 & 6.45 & 5.73 & 1.99 & 5.05 & 4.61 & 1.81 & 6.09 & 5.52 \\
& Linear Probe & 2.89 & 4.20 & 3.98 & 2.23 & 2.64 & 2.58 & 2.35 & 2.50 & 2.48 & 2.42 & 3.07 & 2.98 \\
& Finetune MLP & 2.07 & 2.51 & 2.44 & 2.00 & 2.45 & 2.38 & 1.62 & 2.41 & 2.29 & 1.62 & 2.65 & 2.51 \\
& Full Finetune & 1.71 & 2.38 & 2.27 & 1.70 & 2.32 & 2.23 & 1.60 & 2.30 & 2.20 & 1.68 & 2.36 & 2.27 \\
\bottomrule

\end{tabular}
\caption{Results of two calibration head poses and different numbers of calibration points on Static-MobilePoG. "Same" and "Diff." represent the test errors of the calibrated model on test samples with the same and different head poses as the calibration head pose, respectively. The metric is the Euclidean distance (cm).}
\label{table:two_pose}
\end{table}

\begin{table}[ht]
\centering
\renewcommand\tabcolsep{3.5pt}
\scriptsize
\begin{tabular}{l|c|ccc|ccc|ccc|ccc}
\toprule
\multirow{2}{*}{\textbf{Model}} & \multirow{2}{*}{\textbf{ \makecell{Calibration \\ Algorithm}}} & \multicolumn{3}{c|}{\textbf{1 calibration point}} & \multicolumn{3}{c|}{\textbf{5 calibration points}} & \multicolumn{3}{c|}{\textbf{9 calibration points}} & \multicolumn{3}{c}{\textbf{13 calibration points}} \\
& & Same & Diff. & Avg. & Same & Diff. & Avg. & Same & Diff. & Avg. & Same & Diff. & Avg.\\
\midrule

\multirow{5}{*}{\textbf{iTracker}} 
& w/o calibration & \multicolumn{12}{c}{\cellcolor{gray!20} 2.82} \\
& SVR & 4.17 & 5.99 & 5.54 & 2.37 & 4.47 & 3.98 & 1.74 & 4.02 & 3.52 & 1.77 & 4.21 & 3.71 \\
& Linear Probe & 2.75 & 3.10 & 3.02 & 2.51 & 3.03 & 2.91 & 2.14 & 2.75 & 2.62 & 2.38 & 3.17 & 3.01 \\
& Finetune MLP & 2.59 & 2.63 & 2.62 & 2.11 & 2.85 & 2.68 & 1.75 & 2.73 & 2.51 & 1.81 & 2.77 & 2.58 \\
& Full Finetune & 2.32 & 2.53 & 2.48 & 2.06 & 2.48 & 2.38 & 1.90 & 2.52 & 2.38 & 1.73 & 2.49 & 2.33 \\
\midrule

\multirow{5}{*}{\textbf{AFFNet}}
& w/o calibration & \multicolumn{12}{c}{\cellcolor{gray!20} 2.54} \\
& SVR & 3.73 & 5.90 & 5.36 & 2.27 & 4.41 & 3.91 & 2.03 & 4.19 & 3.72 & 1.84 & 4.08 & 3.63 \\
& Linear Probe & 2.62 & 2.96 & 2.88 & 2.25 & 2.75 & 2.64 & 2.09 & 2.66 & 2.54 & 2.30 & 2.75 & 2.66 \\
& Finetune MLP &2.38 & 2.37 & 2.38 & 1.86 & 2.54 & 2.38 & 1.89 & 2.32 & 2.22 & 1.73 & 2.47 & 2.32 \\
& Full Finetune &  2.05 & 2.31 & 2.25 & 1.64 & 2.28 & 2.13 & 1.57 & 2.21 & 2.07 & 1.59 & 2.23 & 2.11 \\
\bottomrule

\end{tabular}
\caption{Results of three calibration head poses and different numbers of calibration points on Static-MobilePoG. "Same" and "Diff." represent the test errors of the calibrated model on test samples with the same and different head poses as the calibration head pose, respectively. The metric is the Euclidean distance (cm).}
\label{table:three_pose}
\end{table}

\begin{table}[ht]
\centering
\renewcommand\tabcolsep{3.5pt}
\scriptsize
\begin{tabular}{l|c|ccc|ccc|ccc|ccc}
\toprule
\multirow{2}{*}{\textbf{Model}} & \multirow{2}{*}{\textbf{ \makecell{Calibration \\ Algorithm}}} & \multicolumn{3}{c|}{\textbf{1 calibration point}} & \multicolumn{3}{c|}{\textbf{5 calibration points}} & \multicolumn{3}{c|}{\textbf{9 calibration points}} & \multicolumn{3}{c}{\textbf{13 calibration points}} \\
& & Same & Diff. & Avg. & Same & Diff. & Avg. & Same & Diff. & Avg. & Same & Diff. & Avg.\\
\midrule

\multirow{5}{*}{\textbf{iTracker}} 
& w/o calibration & \multicolumn{12}{c}{\cellcolor{gray!20} 2.82} \\
& SVR & 3.83 & 4.62 & 4.36 & 2.13 & 3.45 & 3.04 & 2.09 & 3.15 & 2.84 & 1.85 & 3.10 & 2.76 \\
& Linear Probe & 2.73 & 2.78 & 2.76 & 2.55 & 2.67 & 2.63 & 2.67 & 2.71 & 2.70 & 3.03 & 2.99 & 3.00 \\
& Finetune MLP & 2.64 & 2.65 & 2.64 & 2.25 & 2.62 & 2.51 & 2.10 & 2.63 & 2.47 & 2.03 & 2.56 & 2.42 \\
& Full Finetune & 2.53 & 2.47 & 2.49 & 2.19 & 2.41 & 2.34 & 2.02 & 2.37 & 2.27 & 1.96 & 2.30 & 2.21 \\
\midrule

\multirow{5}{*}{\textbf{AFFNet}}
& w/o calibration & \multicolumn{12}{c}{\cellcolor{gray!20} 2.54} \\
& SVR & 3.57 & 3.87 & 3.79 & 1.97 & 2.84 & 2.57 & 1.77 & 2.79 & 2.49 & 1.74 & 2.77 & 2.48 \\
& Linear Probe & 2.47 & 2.36 & 2.40 & 2.19 & 2.34 & 2.30 & 2.26 & 2.56 & 2.47 & 2.62 & 2.91 & 2.83 \\
& Finetune MLP & 2.39 & 2.24 & 2.29 & 2.07 & 2.30 & 2.23 & 1.97 & 2.32 & 2.22 & 1.92 & 2.31 & 2.20 \\
& Full Finetune & 2.20 & 2.14 & 2.16 & 1.96 & 2.08 & 2.04 & 1.88 & 2.10 & 2.03 & 1.85 & 2.06 & 2.00 \\
\bottomrule

\end{tabular}
\caption{Results of four calibration head poses and different numbers of calibration points on Static-MobilePoG. "Same" and "Diff." represent the test errors of the calibrated model on test samples with the same and different head poses as the calibration head pose, respectively. The metric is the Euclidean distance (cm).}
\label{table:four_pose}
\end{table}

\end{document}